\documentclass{article}

\usepackage{graphicx}%
\usepackage{multirow}%
\usepackage{amsmath,amssymb,amsfonts}%
\usepackage{amsthm}%
\usepackage{mathrsfs}%
\usepackage[title]{appendix}%
\usepackage{textcomp}%
\usepackage{manyfoot}%
\usepackage{booktabs}%
\usepackage{algorithm}%
\usepackage{algorithmicx}%
\usepackage{algpseudocode}%
\usepackage{listings}%

\usepackage[pdftex,dvipsnames]{xcolor}
\usepackage{tabularx}

\usepackage[
backend=biber,
style=ieee,
sorting=none
]{biblatex}
\usepackage[pdftex,dvipsnames]{xcolor}

\usepackage{hyperref}
\hypersetup{
  colorlinks   = true, 
  urlcolor     = blue, 
  linkcolor    = blue, 
  citecolor   = red 
}

\begin{document}

\begin{titlepage}

\vskip 1.5cm

\begin{center}

{\Large \bfseries
AI Foundation Models for Weather and Climate:\\ Applications, Design, and Implementation
}

\vskip 0.5cm

S. Karthik Mukkavilli\textsuperscript{1$*\dagger$}, Daniel Salles Civitarese\textsuperscript{1$\dagger$},
Johannes Schmude\textsuperscript{1$*\dagger$},\\
Johannes Jakubik\textsuperscript{1,2}, Anne Jones\textsuperscript{1}, Nam Nguyen\textsuperscript{1},\\
Christopher Phillips\textsuperscript{3}, Sujit Roy\textsuperscript{3,4}, Shraddha Singh\textsuperscript{1},\\
Campbell Watson\textsuperscript{1}, Raghu Ganti\textsuperscript{1}, Hendrik Hamann\textsuperscript{1},\\
Udaysankar Nair\textsuperscript{3}, Rahul Ramachandran\textsuperscript{4$*$}, Kommy Weldemariam\textsuperscript{1$*$}

\vskip 0.5cm

\textsuperscript{1}IBM Corporation.\\
\textsuperscript{2}Karlsruhe Institute of Technology, Germany.\\
\textsuperscript{3}Atmospheric and Earth Science, University of Alabama, Huntsville, AL, USA.\\
\textsuperscript{4}NASA Marshall Space Flight Center, Huntsville, AL, USA.

\vskip 1.0cm

\textbf{Abstract}
\end{center}

\noindent
Machine learning and deep learning methods have been widely explored in understanding the chaotic behavior of the atmosphere and furthering weather forecasting. There has been increasing interest from technology companies, government institutions, and meteorological agencies in building digital twins of the Earth. Recent approaches using transformers, physics-informed machine learning, and graph neural networks have demonstrated state-of-the-art performance on relatively narrow spatiotemporal scales and specific tasks. With the recent success of generative artificial intelligence (AI) using pre-trained transformers for language modeling and vision with prompt engineering and fine-tuning, we are now moving towards generalizable AI. In particular, we are witnessing the rise of AI foundation models that can perform competitively on multiple domain-specific downstream tasks. Despite this progress, we are still in the nascent stages of a generalizable AI model for global Earth system models, regional climate models, and mesoscale weather models. Here, we review current state-of-the-art AI approaches, primarily from transformer and operator learning literature in the context of meteorology. We provide our perspective on criteria for success towards a family of foundation models for nowcasting and forecasting weather and climate predictions. We also discuss how such models can perform competitively on downstream tasks such as downscaling (super-resolution), identifying conditions conducive to the occurrence of wildfires, and predicting consequential meteorological phenomena across various spatiotemporal scales such as hurricanes and atmospheric rivers. In particular, we examine current AI methodologies and contend they have matured enough to design and implement a weather foundation model.


\Footnotetext{$\dagger$}{Equal contribution.}
\Footnotetext{$*$}{Corresponding authors: karthik.mukkavilli@ibm.com, johannes.schmude@ibm.com, rahul.ramachandran@nasa.gov, kommy@ibm.com.}

\end{titlepage}

\setcounter{tocdepth}{2}
\tableofcontents

\section{Introduction and Overview}
\label{sec:introduction}

Powered by the availability of large datasets for training and rapid improvements in GPU-driven computation, the 2010s saw rapid progress and various breakthroughs in deep learning. Following this revolution, research efforts originating in natural language processing (NLP) with large language models (LLM) using transformers and self-supervised learning have led to the emergence of the \emph{foundation model} paradigm \cite{devlin2018bert,brown2020language,radford2021learning}. As defined in \cite{Bommasani2021}, a foundation model is a model trained on large data volumes in a self-supervised (i.e., task-independent) manner and can be \emph{effectively fine-tuned} to several downstream tasks. Typically, foundation models are large, consisting of several million or even billions of parameters.

Large language models have recently benefited most from these efforts, but the weather and climate domains also possess experience with massive data volumes, super-computing, and machine learning. Moreover, despite numerical weather prediction's (NWP) great success in modeling the physical processes of the atmosphere \cite{bauer2015quiet}, data-driven machine learning methods to post-process NWP outputs as model output statistics have been used as early as the 1970s \cite{glahn1972use}. Additionally, the volume of gridded weather and climate model outputs is growing daily, with meteorological, climate services and research labs of all major nations contributing to this growth. European Centre for Medium-Range Weather Forecasts (ECMWF) archives contain about 450 PB of data to which 300 TB are added daily \cite{ecmwf_webpage}. This paper specifically reviews development of foundation models utilizing such datasets to address weather and climate prediction problems. We are interested in applications ranging from detecting and forecasting extreme weather events (hurricanes, atmospheric rivers, fire weather conditions) and their impacts to hybridizing AI with NWP.

Adopting the foundation model paradigm can result in various advantages. Many applications experience model performance and accuracy breakthroughs, and other benefits (e.g., data efficiency) are also possible. Indeed, for many use cases where data are limited (whether labels or covariates), the foundation model paradigm provides a clear path toward developing a robust and efficacious model \cite{devlin2018bert, brown2020language, dosovitskiy2020an, liu2021swin, liu2023video}. This is due to the extensive self-supervised pre-training, which reduces the volume of labeled data required for fine-tuning specific tasks \cite{hendrycks2020pretrained, zhang2022delving}. Finally, even if data are abundant, training large models is expensive. In the course of this review, we will encounter models that have been trained on dozens of GPUs for weeks or even months. Instead of committing to such cost per use case, following the foundation model approach means that large-scale training is done only once since fine-tuning can be performed on a handful of GPUs in hours or days. State-of-the-art AI models that are competitive against NWP models are emerging \cite{lam2022graphcast, pathak2022fourcastnet, bi2023accurate, chen2023swinrdm}. However, these models currently are limited by physical scales, suffer from long-term rollout, and do not generalize to multiple downstream tasks. 

Foundation models typically consist of an encoder $f_\theta$ and a decoder $g_\phi$, where $\theta$, $\phi$ stands schematically for parameters. Pre-training the foundation model then optimizes
\begin{equation}\label{eq:encoder_decoder_pattern}
 g_\phi \circ f_\theta
\end{equation}
using some self-supervised optimization strategy. However, when fine-tuning the model for downstream tasks, one keeps $f_\theta$ but discards the decoder $g_\phi$ and replaces it with a task-specific decoder $h_\psi$. That is, fine-tuning optimizes $h_\psi \circ f_\theta$. Typically, decoders are very lightweight, while encoders can comprise hundreds of millions or even billions of parameters because they need to understand and represent the complexity of the input data. Therefore, foundation models reduce training overhead by only using task-specific decoders. To understand why this process works, one can consider the encoder as learning the dynamics and relationships that exist within the data during pre-training, while fine-tuning teaches the decoder to apply these relationships to a specific task. Of course, this illustration is limited and only aids conceptual understanding. Despite this, it should be clear that foundation models are heavily rooted in self-supervised and representation learning. Similarly, it should also be clear that the data used for pre-training needs to be such that the learned distribution can represent, at some level, all the variables and the underlying physics necessary to implement models for all downstream tasks.

Although a wealth of data is available and numerous applications exist in the fields of weather and climate, there still exists a need to explore self-supervised learning further in these contexts. Recently, there have been significant developments in massive models for weather forecasting, but these models primarily rely on supervised learning. Nevertheless, there are examples of representation learning in this field, such as those found in \cite{hoffmann2023atmodist,lessig2023atmorep,nathaniel2023resource}. Currently, the only publicly available foundation models for weather and climate are \cite{nguyen2023climax,lessig2023atmorep}.

The remainder of this paper is organized as follows. Section \ref{sec:applications} discusses various weather or meteorological applications, giving special attention to the applicability of both machine and deep learning. Given this set of potential downstream tasks, Section \ref{sec:design} concerns the choices one must make when building a foundation model. While it is not unrealistic to expect the ultimate emergence of several foundation models that can model all aspects of earth systems, the current field strongly suggests that one should decide on a range of phenomena and possible applications to target. Setting the scope of a new foundation model will lead to some concrete requirements for its implementation, which is discussed in Section \ref{sec:implementation}; i.e., there, we will discuss the tradeoffs between various AI backbones such as transformers and graph neural networks.

This paper is written with several audiences in mind. AI, weather, and climate researchers and domain experts are interested in specific applications. Some might want to read Sections \ref{sec:applications} and \ref{sec:design} while skipping the technical details of deep learning models in Section \ref{sec:implementation}. AI researchers, on the other hand, could skim \ref{sec:applications} and focus on Sections \ref{sec:design} and \ref{sec:implementation}.

\section{Applications}
\label{sec:applications}
\quad We will examine the potential applications to start our discussion of foundation models for weather and climate. Here, we focus on applications in the sense of possible downstream machine learning tasks in the weather and climate domain, including applications that combine weather/climate information with other data (for example, to predict impacts). We review these to understand the advantages and disadvantages of current ML models, and the potential benefit to each application in using an FM approach, and any design consequences for the FM.

It is important to note that while this is not an exhaustive list, we draw from decades of experience to focus on a select few applications that hold significance for both the scientific and business communities. Briefly, this section is ordered in three parts: Weather forecasting (Section \ref{sec:applications:forecasting}), improving NWP models and their outputs (Sections \ref{sec:applications:model_blending}, \ref{sec:applications:parametrization}, \ref{sec:applications:data-assimilation}, \ref{sec:applications:downscaling}), detection of weather patterns (Section \ref{sec:applications:detection}, modeling weather/climate impacts and other weather-driven applications (Section \ref{sec:applications:impacts}), and applications in climate science (Section \ref{sec:applications:climate}).

\subsection{Forecasting}
\label{sec:applications:forecasting}

Forecasting is among the first problems that come to mind when considering machine learning applications for weather and/or climate. From an AI researcher's point of view, this is likely because the most straightforward issues in AI are frequently those where both labels and established benchmark scores are abundant, as is the case in NWP. However, even if one ignores that there is no single score that accurately reflects the performance of a given weather forecast, one has to keep in mind that the benchmarks are not other AI models but highly complex simulation systems that are the result of decades of research operating on often massive and expensive high-performance computing (HPC) systems \cite{bauer2015quiet}.

Weather forecasting requires different approaches and models at different prediction timescales. Establishing current weather conditions from recent observations or forecasting a few hours into the future is known as \emph{nowcasting}, which generally utilizes a numerical atmospheric model assimilating observations. Regional NWP models are used to forecast a few days into the future. Examples are the High-Resolution Rapid Refresh (HRRR) \cite{dowell2022high} and the High-Resolution Deterministic Prediction System (HRDPS) \cite{milbrandt2016pan} which cover different parts of North America and make predictions at 3 and 2.5 km resolution, respectively, up to 48 hours ahead. Models with global coverage include ECMWF's High-Resolution model (HRES), with forecasts of ten days at 0.1 degrees ($\sim 10$ km) and NCEP's Global Forecast System (GFS), with forecasts of 16 days at 13 km ($\sim 0.125$ degrees). Due to higher uncertainty, sub-seasonal and seasonal forecast models rely more on ensemble members for longer time ranges (weeks to months into the future). Ensemble members are generated by running the model multiple times with perturbed initial conditions. For example, the Climate Forecast System Version 2 (CFSv2) comprises four ensemble members \cite{saha2014ncep}. CFSv2 is a fully coupled model representing the interaction between the Earth's atmosphere, oceans, land, and sea ice. Also of interest are atmospheric reanalysis datasets, where a numerical model is used to assimilate observations and create a long-term, consistent best estimate of the earth-atmosphere system state.

As we also discuss later in Section \ref{sec:design:scales}, the choice of prediction time scale has potentially profound implications for modeling approaches and design. At the shortest time scales, conventional NWP struggles to provide useful lead times, and performant nowcasts of precipitation from radar imagery can be made using what is known as Eulerian and Lagrangian persistence \cite{germann2002scale}; that is, without tackling the full complexities of atmospheric processes, let alone interactions between the atmosphere and Earth's surface.

With all this in mind, it should be no surprise that the first successes in forecasting using AI models focused on nowcasting. Here, techniques from computer vision, such as optical flow and relatively simple deep learning models, allow the propagation of a meteorological pattern (defined by an image) a few pixels into the future without tackling the full complexity of the underlying system
\cite{ayzel2019optical,agrawal2019machine,sonderby2020metnet,prudden2020review,ravuri_skilful_2021,espeholt2022deep,zhang_skilful_2023,andrychowicz2023deep}.

While the physical system and the vast amount of data involved may seem complex, there is a compelling argument for the potential of AI in addressing forecasting challenges. Operational forecasting NWP models assimilate observations and prior forecasts by solving a series of differential equations while optimizing a cost function to generate new initial conditions for continuous weather forecasting. This process can essentially be distilled into a simple function, denoted as $x_{t+1} = f(x_t)$. Given that neural networks have the capability to approximate nearly any well-behaved function, a suitably sophisticated model should have the capacity to handle, at the very least, the task of projecting a state $x_t$ into the future, provided it has access to a sufficiently diverse set of training data. (The arguably more complex data assimilation problem is discussed in Section \ref{sec:applications:data-assimilation}.)

And indeed, the last few years have seen the emergence of AI emulators of NWP systems \cite{pathak2022fourcastnet, lam2022graphcast, bi2023accurate, bonev2023spherical}. Very large deep learning systems take the state of the atmosphere at time $t$ as input and propagate it one step into the future. Typically, these deep learning systems are trained on re-analysis data, with $x_t$ being a subset of ERA5 parameters \cite{hersbach2020era5}. Technically speaking, these models do not learn to simulate atmospheric processes. They are trained on the gridded output of a re-analysis system and learn to emulate model fluid dynamics and other physical processes. Thus, an optimal performing AI emulator trained with, e.g., RMSE loss on ERA5 data will reproduce ERA5 data along with all its biases.

The field of AI forecasting has made significant progress, as seen in the notable work of \cite{pathak2022fourcastnet, lam2022graphcast, bi2023accurate, bonev2023spherical}, which has garnered attention from the meteorological community. While some researchers argue that their models outperform established NWP systems like ECMWF's Integrated Forecast System (IFS), also known as HRES, we can confidently state that these models perform comparably to IFS and similar systems. Evaluating weather forecasts is a complex task, and AI emulators have several advantages that make them appealing. The most significant is their speed. Despite requiring several weeks or months of training on multiple GPUs, AI emulators can make predictions using just one or a few GPUs in seconds. Such an advantage makes AI interpolation three orders of magnitude faster than NWP predictions on HPC systems.

However, as with any emerging research area, all existing AI emulators have clear limitations. First, the spatial resolution is limited to the training data, which for all of the above models is ERA5 data due to the availability of a long time series of global coverage (1940 to Present) and good spatial resolution
(25km). Consequently, these models exhibit, at best, a resolution of 0.25 degrees compared to the 0.1 degrees of HRES, let alone the kilometer-scale resolutions of regional models like HRRR and HRDPS. (See Section \ref{sec:applications:downscaling} for potential downscaling applications.)

In addition, AI emulators frequently only model a subset of parameters, time steps, and vertical levels. The reason for that is simply one of scales. ERA5 data is distributed on 37 pressure levels \footnote{While NWP often uses terrain-following vertical coordinate system, meteorological fields interpolated to pressure levels are better suited for analysis by forecasters. For the purpose of our discussion, it is sufficient to be aware that NWP model outputs for general use is often distributed on fixed pressure levels.}, representing the six primary parameters\footnote{Easterly, northerly and vertical wind as well as temperature, humidity and geopotential.} at each grid point, which means committing about 900 MB of memory at 32-bit resolution. Typically, this is addressed by curating a subset of pressure levels, time steps, and parameters. Which leads to the limitation mentioned above.

Moving beyond issues of richness in spatial or temporal resolution as well as numbers of parameters, AI emulators currently exhibit poor performance at longer lead times. The predictions become increasingly ``blurred'' as lead times increase, and they fail to capture extremes in the same manner as ensemble NWP forecasts. This is, among other things, thought to be due to the RMSE training criterion, an issue we will return to in Section \ref{sec:design:rollout}. Finally, some emulators are trained such that different lead times are treated separately, which can lead to temporal inconsistencies.

As mentioned above, datasets such as ERA5 are created using NWP models assimilating observations to produce physically consistent best estimates of atmospheric state. Thus, such datasets incorporate both observational and model biases \footnote{We are making this distinction since the observational system can introduce biases.}. However, note there are deep learning models that directly use observations, as we mentioned earlier in the context of nowcasting and shorter lead times. One example is learning-based nowcasting of precipitation using radar data \cite{ravuri_skilful_2021, zhang_skilful_2023}, extended by the authors of \cite{espeholt2022deep} through fusing this data with NWP analyses. Finally, \cite{andrychowicz2023deep} includes further weather variables in a forecast using sparse observations. These models operate on spatial scales equal to or finer than convection-resolving NWP models, although the latter \cite{andrychowicz2023deep} combines dense gridded data with sparse point data from weather stations. Further discussion on climate models is in \ref{sec:applications:climate}.

\subsection{Model Blending and Post-processing}
\label{sec:applications:model_blending}

While skillful AI-based forecasting is a more recent development, there is a long history of approaches that attempt to achieve superior performance by applying data-driven techniques to the output of NWP systems. Effectively, one can think of these as reasonably lightweight approaches that sit on the shoulders of giants.

 Various model output statistics have been implemented that date back to the 1970s \cite{glahn1972use}. These statistics consider the model's predictions and other covariates as input and then trains a machine-learning model to predict observations. A direct extension of this approach is that of model blending, where the model considers the output of multiple NWP models. See, e.g., \cite{lu2015machine} for further details.

Whether one considers a single NWP model or several, these models frequently exhibit better performance scores than their inputs. Since the ML models are blissfully unaware of the various constraints encoded in the NWP models to make predictions, they are free to violate them in the pursuit of optimization (although techniques do exist to enforce hard constraints if they can be appropriately defined; see \cite{harder2022constraints}). When things go well, the models are thus able to correct model biases. At the same time, these models are free to make predictions that are incompatible with atmospheric dynamics and physics.

\subsection{Downscaling}
\label{sec:applications:downscaling}

In the context of our discussion of AI emulators in Section \ref{sec:applications:forecasting}, we pointed out that current emulators were restricted in terms of their spatial resolution and training data. One way to address this is by \emph{downscaling}, also called super-resolution. Here, the problem is increasing the resolution of some gridded weather data beyond what is possible by naive interpolation. This task is challenging in part due to differences in implementation between different NWP models.  Because, e.g., HRRR and ERA 5 assimilate disparate observations and implement the atmosphere's governing equations differently, it is not sufficient to train an AI to predict the HRRR from ERA 5. Such a task requires not only learning the super-resolution relations, but also transforming one set of model biases and errors to another. Super-resolution training datasets must be created that are consistent in their source data, typically by interpolating a single NWP model to other grids.

The most basic deep learning approaches use convolutional neural networks, frequently U-Nets \cite{ronneberger2015u}, to generate high-resolution outputs from low-resolution inputs. Later on, Generative Adversarial Networks (GANs) were discovered to be a good avenue to boost the representation of finer structures and details \cite{stengel2020adversarial,wang2021fast}. More recent work leverages AI concepts as diverse as normalizing flows \cite{groenke2020climalign} and neural operators \cite{yang2023fourier}. The authors of \cite{hoffmann2023atmodist} used representations obtained by self-supervised learning for downscaling, an avenue that was also explored in \cite{nguyen2023climax}.

\subsection{Parameterization}
\label{sec:applications:parametrization}

 In the preceding two sections, our discussion focused on improving forecast models. This was achieved by increasing resolution or by placing outputs in the context of observations. In this short section and the subsequent section \ref{sec:applications:data-assimilation}, we will conclude our discussion on how to improve NWP models by discussing how AI might be part of the models themselves. This contrasts our discussion, where NWP models were largely treated as black boxes, with inputs and outputs fed into AI systems.   

 A key area of research that means to augment and support or accelerate NWP systems with AI models is that of parameterization \cite{pielke2006new}. Parameterization is where algorithmic or statistical methodologies are used to represent physical processes, where sub-grid scale processes, complexity or lack of understating prevents explicit representation in a model \cite{warner_2010}. Given that it does not always rely on first principles, NWP parameterization is an area ripe for data-driven approaches. For example, \cite{rasp_deep_2018} used a deep learning emulator trained on a high-resolution cloud-resolving simulation model to replace the parameterization of clouds in a climate model. Additionally, benchmark data sets such as ClimSim are being developed to aid research into this area \cite{YuClimSim}. For more recent work see \cite{yuval2020stable,brajard2021combining,lutjens2022multiscale}.

\subsection{Data Assimilation}
\label{sec:applications:data-assimilation}

As alluded to earlier, a fully-fledged NWP system not only propagates a state into the future; it also assimilates observational data to establish the initial state and to constrain the evolution of the model to future states. This process is known as data assimilation (DA). The DA process is complex, costly, and thus an area of ongoing research. In addition, the prevalent approaches rely on simplifying assumptions such as linearity. While a challenging field, there is increasingly promising research to leverage deep learning in this context. See \cite{mack2020attention,arcucci2021deep,farchi2022online}.

\subsection{Detection and Prediction of Weather Patterns}
\label{sec:applications:detection}

Automated detection of extreme weather features in reanalysis data and climate projections can help to quantify better the risk of extreme climate impacts from storms and compound hazards in future climates \cite{dowdy_extreme_2017, catto_understanding_2021}. Traditional approaches have been to use systematic detection algorithms with empirical heuristics for features such as fronts, tropical cyclones (TCs), extra-tropical cyclones (ETCs), and atmospheric rivers (ARs) in reanalysis or climate model projection data (e.g., \cite{prabhat_teca_2015}). Additionally, specialized systems for predicting specific features such as tornadoes or hurricanes may offer advantages over general reanalysis-trained models that predict the entire atmospheric state, such as access to high-resolution observations like radar.

More recently, applications of deep learning for weather feature detection have been explored to eventually provide a more general solution and avoid the heuristics methods' reliance on specific variables and thresholds, which are often dataset-specific. \cite{prabhat_climatenet_2021} presents an expert-labeled dataset of tropical cyclones (TCs) in the 25km CAM5.1 model and demonstrates rapid segmentation of TCs and ARs using an architecture developed for spatial segmentation in computer vision applications. \cite{racah_extremeweather_2017} implements the detection and localization of TCs, ETCs, ARs, and tropical depressions. While \cite{racah_extremeweather_2017} use a 3D convolutional model with time as one of the dimensions, \cite{prabhat_climatenet_2021} train on static 2D images but report reasonable temporal consistency in model predictions. Other examples of weather feature detection are tornado diagnosis in convection-resolving dynamic forecasts using CNNs and logistic regression \cite{sobash_diagnosing_2023} and diagnosis of turbulent conditions in regional or high-resolution weather forecasts \cite{munozAviationTurbulence}. In the latter application, the authors used pre-computed turbulence indices in combination with traditional ML methods (e.g., XGBoost), so there is potential for improvement using more sophisticated ML methods capable of unsupervised feature extraction.

In addition to simply detecting weather features, AI models have also been applied to predicting their evolution. For example, nowcasting of tornadoes has been accomplished using CNNs on composite radar \cite{lagerquist_deep_2020} and multi-modal data \cite{barajas_performance_2019}. Hurricane or tropical cyclone (TC) forecasting primarily focuses on predicting the storm's trajectory and intensity (e.g., max wind speed). ML has been used to model each of these separately or in combination as part of a meta multimodel framework \cite{boussioux_hurricane_2022}. Datasets consumed by such models consist of reanalysis data supplemented by TC observations and operational hurricane forecasts (comprising both statistical and dynamical models, including resolutions up to about 3km). Any downstream task concerned with hurricane or TC forecasting would, therefore, need to consume higher resolution dynamical model data due to the poor representation of TCs in reanalysis datasets \cite{bian_how_2021}. Cited challenges again include the small number of extreme events for supervised learning and the need to integrate a broader range of datasets.

One challenge of training supervised detection models is the limited quantity of labeled data, particularly for extreme events. The task-independent foundation model approach may reduce the challenge of limited labeled data compared to a task-specific supervised deep learning model. We also note that previous studies have tended to train and test on outputs from single models and at fixed resolution. A multi-dataset, multi-scale foundation model approach may also enable better generalization (whereby fine-tuning can learn the particular biases and variables of importance for specific datasets). 

One question to be resolved is whether a foundation model designed to enable forecast emulation as a downstream task would be broad enough to enable weather feature segmentation, such as atmospheric river identification. We also note that weather/climate feature detection is potentially valuable across prediction timescales simply by applying the model to NWP or climate model projection data as required.

\subsection{Incorporating ML in Climate Science}
\label{sec:applications:climate}

Machine learning (ML) is becoming a pivotal tool in climate science, enhancing our comprehension of climate systems and facilitating faster and more accurate climate projections. ML techniques are instrumental in analyzing extensive climate data sets to identify patterns and trends, improving the precision of climate models, projecting potential climate changes under various emission scenarios, and evaluating the impacts of climate change on diverse regions and sectors. Recent publications in CMIP projects, including works by \cite{schneider2023climate,YuClimSim,Molina2023}, underscore the growing integration of ML in addressing various facets of climate science, from uncertainty quantification, speeding up simulations via ML based paramaterizations, designing emulators, development of datasets and tools for hybrid ML-physics research, uncertainty quantification to using ML to identify the sources of predictability for modes of climate variability. 

Despite the promising prospects, employing ML in climate science is not devoid of challenges. The intricacies of climate systems and the often limited or incomplete climate data pose significant hurdles in developing ML models that can aptly represent the climate system's behavior. Additionally, the inherent uncertainty in climate projections remains a critical limitation, although ML can somewhat alleviate this by reducing the uncertainties linked with emission scenarios \cite{rolnick2022tackling}. 

Nevertheless, with ongoing advancements, ML promises to further amplify its role in climate science. As ML techniques continue to evolve and with the anticipated availability of more comprehensive climate data, ML is set to cement its significance in enhancing our understanding of climate change and making more precise future climate change projections.

The rich repository of climate data generated by the CMIP6 project serves as a vital resource in augmenting the efficacy of ML models in various domains, including predicting forthcoming climatic alterations, recognizing extreme weather occurrences, and evaluating the repercussions of climate change. Utilizing CMIP6 as a benchmark, the performance of ML models can be meticulously assessed, fostering continuous improvements and innovations. Furthermore, CMIP6 facilitates synthetic data creation, a crucial asset in training ML models where actual data is scant or unavailable. This symbiosis also sparks the development of novel ML techniques, enhancing the capacity to comprehend and forecast intricate systems like the climate. This integration signifies a leap towards harnessing computational intelligence in augmenting our understanding of climate dynamics and formulating data-driven strategies for climate resilience. Ultimately, this can inform climate adaptation strategies \cite{jain2023ai} for impact assessments \cite{jones2023ai} and solutions to tackle climate change \cite{rolnick2022tackling}.

\subsection{ML Impact and Application Models}
\label{sec:applications:impacts}

Developing machine learning models for impacts and applications of weather and climate involves translating weather and climate data into variables suitable for the specific application system, that while responsive to the atmosphere, are not themselves atmospheric variables. This may also include integrating additional non-climate drivers and related datasets. The complexity of such systems can vary significantly, resulting in variations in the intricacy of subsequent modeling tasks and influencing the potential benefits and relevance of incorporating pre-trained weights from foundational models for these specific applications.

Modeling complex hazards such as floods and wildfires typically involves using specific complex numerical models with multivariate inputs. For example, ML is well established as a tool in flood forecasting but is generally used as one or more components of a multi-stage hybrid modeling framework (e.g., \cite{nevo_flood_2022}) or as an in-situ surrogate for hydrological simulations \cite{guo_data-driven_2021}. Modeling a process such as flooding may require a separate FM in itself. These applications are, therefore, not apparent candidates for downstream FM tasks, although they may still benefit from inputs from FM-derived forecasts or downscaled data.

ML has been used extensively for solar radiation, wind, and power output prediction, with timescales mainly focused on minutes to hours ahead\cite{alkhayat_review_2021}. However, solar forecasting with mesoscale high-resolution models, e.g., NCAR's weather research and forecasting (WRF) models, where cloud evolution is usually beyond 6 hours and requires energy dispatch to capture peak electricity prices \cite{MesoscaleSimulations}, could likely be accelerated with machine learning surrogates of atmospheric dynamics and radiative transfer. Moreover, solar resource assessment from variability in seasonal and climate variables, such as aerosols and dust \cite{Mukkavilli}, could benefit from machine learning-based approaches \cite{hoyne2019deep}. With limited sun photometers and aerosol observational networks, pre-training with a foundation modeling can enable inference with limited labels.

\subsection{Benefits of Foundation Modeling for Weather and Climate Applications}
\label{sec:applications:summary}

Foundation modeling is emerging as a pivotal tool with the potential to significantly improve the efficacy and precision of weather and climate applications. Below, we outline potential benefits that we believe are important for weather and climate domain:

\begin{enumerate}
 \item \textbf{Increased Accuracy and Efficiency}: Foundation models offer both increased accuracy and computational efficiency in forecasting atmospheric variables across diverse time scales. We believe that this provides a significant competitive edge when compared to traditional numerical models.
 
 \item \textbf{Mitigation of Supervision Bottleneck}: By virtue of being pre-trained on extensive datasets, these models enhance their robustness by mitigating the supervision bottleneck. This reduction in reliance on supervised learning not only decreases the demand for large quantities of labeled data but also enhances their performance in real-world deployments.

 \item \textbf{Versatility and Adaptability}: Foundation models can be easily fine-tuned to address diverse tasks, encompassing weather prediction, climate modeling, hazard analysis, and more. They can handle varied scenarios and situations in Earth's atmospheric studies, offering valuable versatility. What's particularly advantageous is that they require significantly less labeled data and reduced additional supervision compared to traditional ML models. This adaptability and versatility make foundation models valuable for researchers and practitioners in Earth's atmospheric studies, simplifying applying AI-powered solutions to complex challenges in this domain.
 
 \item \textbf{Improved Generalization}: Exhibiting superior generalization capabilities, these models can manage shifts outside their training distribution, e.g., to enhance their reliability in forecasting extreme weather events and long-term climate projections.

\item \textbf{Rich and Multimodal Data Utilization and Handling}: Foundation models can integrate physics-informed climate simulation models for pre-training, offering a rich source of data that augments the learning process and enables more informed predictions. They proficiently handle the highly multimodal nature of climate data by integrating varied data types and irregular datasets through innovative techniques, thereby enhancing model performance.

 \item \textbf{Scalability and Innovations}: Exhibiting promising scalability, the foundation models can provide enhancements with larger, higher-resolution datasets, heralding a new generation of multi-scale and multi-physics data-driven models in Earth system science, along with recent innovations like FourCastNet \cite{kurth2023fourcastnet}, Panguweather \cite{bi2023accurate}, and GraphCast \cite{lam2022graphcast} that depict rapid advancements in this field.

 \item \textbf{Broader Impact and Potential}: Integrating existing climate science knowledge with machine learning models, it is possible to achieve faster inference and data assimilation capabilities, potentially creating a broader impact on climate science and policy-making, inclusive of facilitating government planning and disaster mitigation efforts of societal relevance.

 \item \textbf{Potential for Integration with Multi-Sensory Data}: The emerging trend of foundation models hints at the prospective integration with multi-sensory data in weather and climate science, promising to broaden the scope and accuracy of future predictions and analyses.
\end{enumerate}

\section{Design}
\label{sec:design}

Designing a foundation model (FM) for weather and climate requires considering the relevant scales discussed in Section \ref{sec:applications}. The Earth system operates at various temporal scales, from seconds to centuries, and encompasses a diverse spatial range from micro to global planetary scales. It is also important to consider external cycles associated with our solar system that impact Earth's weather. Therefore, identifying areas where machine learning can add the most value through foundational modeling and considering downstream tasks that can be feasibly fine-tuned concerning societal and business implications is crucial. Likewise, we need to provide enough pre-training data showing sufficient richness in phaenomena to cover all requirements for downstream tasks. Typically, this means more extended time series covering decades, for example, ERA5, MERRA2, or even CMIP6. As we look to the future, we may have individual FM systems for Earth's subsystems: the atmosphere, hydrosphere, cryosphere, biosphere, and geosphere.

\begin{figure}[ht]
\vskip 0.1in
\begin{center}
\centerline{\includegraphics[width=1.0\textwidth]{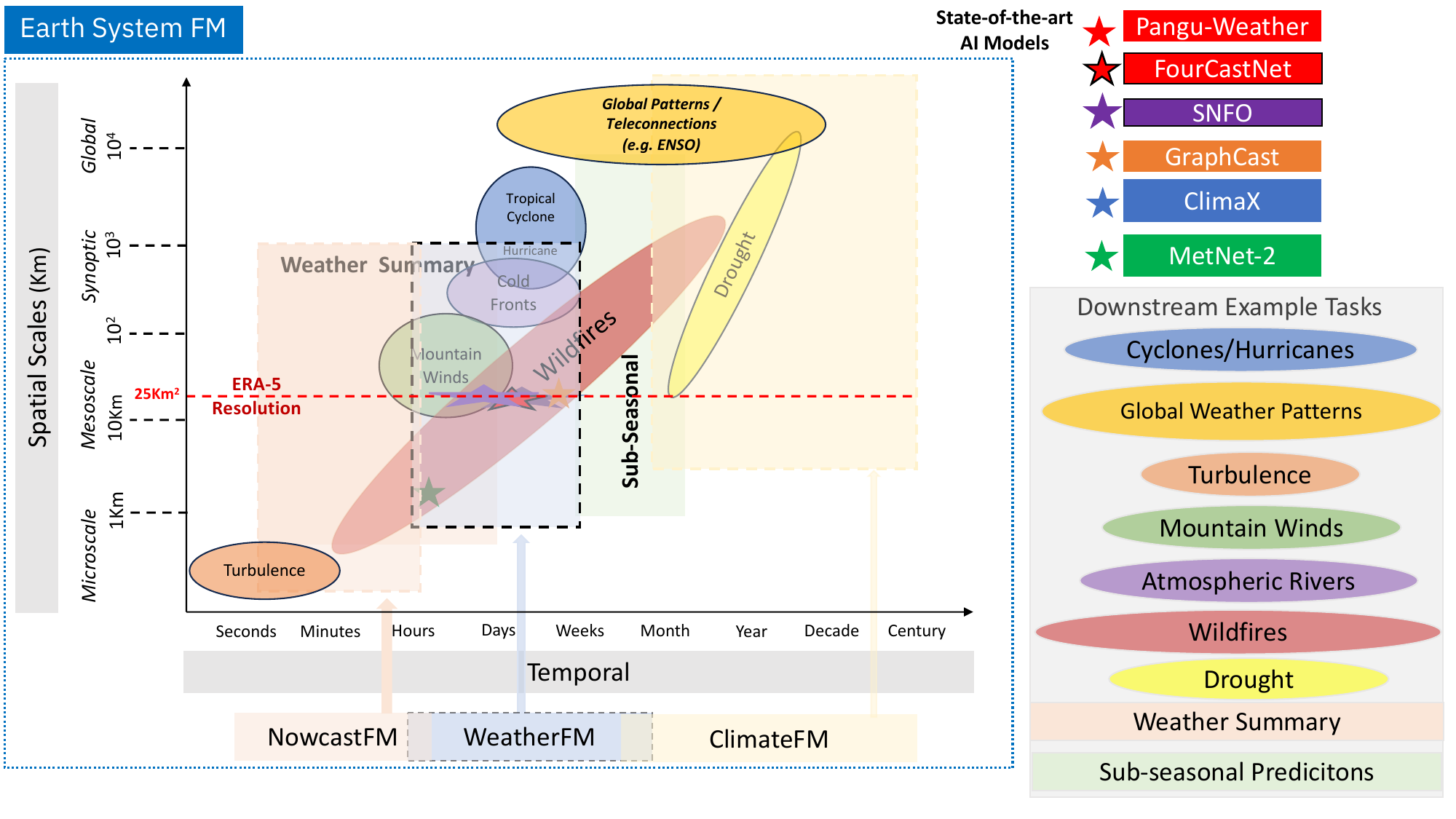}}
\caption{Meteorology and climate science perspective on foundation models, with spatiotemporal windows of focus. WeatherFM is the most mature, but as indicated, current state-of-the-art AI models using operator learning, transformers, and graphs are mostly around the red ERA5 spatiotemporal line within the broader window. Scopes of FM: Earth System (blue dots/box); nowcasting (orange dash/box); weather (black dash, blue box); climate (yellow dash/box).}
\label{fig:WeatherFM_meteo}
\end{center}
\vskip -0.2in
\end{figure}

During Section \ref{sec:applications:forecasting}, we delved into the challenges of obtaining high spatiotemporal resolution datasets like ERA5 and MERRA-2 globally, compared to HRRR, which is only available for CONUS. CMIP6 offers an excellent chance for pretraining large-scale models, but handling inconsistent variables across multiple sources is challenging. Consequently, the most technologically advanced approach for the community is to model temporal scales that span from hours to weeks within 10-14 days (less than sub-seasonal). Figure \ref{fig:WeatherFM_meteo} illustrates the intricate multi-scale complexity of Earth system modeling. To start our discussion, we examine the models presented in \cite{bi2023accurate, cachay2023dyffusion, bonev2023spherical, ruhe2023geometric,mialon2023self, raonic2023convolutional} to identify their design criteria.

Pangu-Weather \cite{bi2023accurate} introduces a novel AI-based method for medium-range global weather forecasting, which performs competitively against traditional NWP methods on accuracy and significantly outperforms NWP on computational efficiency for inference. The key innovation is the 3D Earth-specific transformer (3DEST), a deep learning architecture designed to handle the three-dimensional nature of atmospheric data. By formulating height as an individual dimension, the 3D models effectively capture complex patterns in weather data and the relationships between atmospheric states at different pressure levels. The model's Earth-specific positional bias enhances its ability to handle Earth-specific features in weather data. The hierarchical temporal aggregation strategy also reduces the number of iterations needed for medium-range weather forecasting, resulting in faster and more accurate forecasts. Pangu-Weather demonstrates robust deterministic forecast results that are competitive with the performance of operational NWP systems. Integrating 3D deep neural networks, Earth-specific priors, and hierarchical temporal aggregation has enabled Pangu-Weather to achieve accurate global weather forecasting while significantly reducing computational costs.

The approach in \cite{cachay2023dyffusion} leverages the temporal dynamics encoded in the data to train diffusion models for dynamical system forecasting. The method achieves multi-step and long-range forecasting capabilities by directly coupling the temporal dynamics with the diffusion steps in the network. This design choice enhances flexibility and continuous-time sampling trajectories and allows for trading off performance with accelerated sampling during inference. Moreover, the dynamics-informed diffusion process improves computational efficiency compared to traditional Gaussian noise-based diffusion models. The approach demonstrates competitive performance in complex dynamics forecasting of various systems, including sea surface temperatures, Navier-Stokes flows, and spring mesh systems. This highlights the effectiveness of incorporating atmospheric dynamics into the diffusion model to improve forecasting accuracy and efficiency.

Geometric Clifford Algebra Networks (GCANs) \cite{ruhe2023geometric} aims to model dynamical systems by leveraging symmetry group transformations using geometric algebras. The method introduces group action layers that linearly combine object transformations using pre-specified group actions by encoding isometries as $Pin(p,q,r)$ group elements. These layers and a novel activation and normalization scheme are adjustable geometric templates that can be refined through gradient descent. The approach shows significant advantages in modeling three-dimensional rigid body transformations and large-scale fluid dynamics simulations, outperforming traditional methods. By incorporating geometric algebra and symmetry group transformations, GCANs offer a promising way to encode atmospheric dynamics and model dynamics, enabling improved performance in weather forecasting from a meteorological perspective. GCANs can outperform traditional models of complex dynamical systems, such as three-dimensional rigid body transformations and large-scale fluid dynamics simulations. They leverage symmetry group transformations and geometric algebras, providing theoretical advantages and improved performance. Introducing group action layers as adjustable geometric templates enhances their flexibility and expressiveness in representing dynamics. However, despite the promise of Clifford algebra, the work is in the nascent stages, not yet ready for real-world operational systems without further research.

The approach in \cite{mialon2023self} uses self-supervised learning (SSL) and joint embedding methods to learn general-purpose representations of partial differential equations (PDEs) from heterogeneous data. The method can extract useful information from various sources using SSL, including messy or incomplete observations of real dynamical systems. The learned representations outperform baseline approaches in invariant tasks, such as regressing PDE coefficients, and improve neural solvers' time-stepping performance. This methodology shows promise in developing general-purpose foundational AI models for PDEs, which could have broad impacts in science and engineering, including weather forecasting, by providing computationally efficient alternatives to numerical solvers.

Recent literature on transformer-based architectures for weather forecasting proposed a couple of ways to develop and integrate attention mechanisms. Approaches that have been successfully applied to the weather domain can mainly be divided into 3D self-attention \cite{bi2023accurate} and 2D cross-attention \cite{nguyen2023climax}.
These approaches are particularly successful since they aim at reducing the computational costs associated with the attention mechanism in this data-intense domain. Computational effort can be reduced by employing window-based attention (Swin transformer \cite{liu2021swin}) or by using variable aggregation and cross-attention across spatial positions \cite{nguyen2023climax}. For the window-based approach, feature maps are partitioned into windows, and the attention mechanism (i.e., self-attention) is applied to each window. This approach has successfully been utilized in non-spherical \cite{bi2023accurate} and spherical \cite{carlsson2023heal} settings and has generally demonstrated promising performances. 
Variable aggregation and cross-attention were utilized to reduce the complexity of the attention mechanisms of the plain vision transformer (ViT) architecture. 
It is worth mentioning that besides their computational complexity, attention layers infuse a high model accuracy, capture long-term relationships, can deal with different types of data, can include different kinds of added biases, are parallel by design, and mainly do not require Fourier transform, which allows models to deal with high-resolution data \cite{carlsson2023heal}.

Spherical Fourier Neural Operators (SFNOs) is a generalization of Fourier Neural Operators (FNOs) to model atmospheric dynamics on spherical geometries \cite{bonev2023spherical}. FNOs have successfully learned operators efficiently for spatio-temporal data due to their ability to capture long-range dependencies. However, they suffer limitations in spherical coordinates, leading to visual artifacts and dissipation. SFNOs address these issues by using generalized Fourier transform over compact Riemannian manifolds, in this case, dynamical system over the sphere, and demonstrate stable auto-regressive rollouts for atmospheric dynamics. Leveraging SFNOs in pre-training foundational AI models from a meteorological perspective can enhance the representation of atmospheric dynamics and model dynamics, leading to improved weather forecasting and climate simulations. This advancement could accelerate the development of climate models and aid in understanding and responding to climate change more effectively.
 
Convolutional Neural Operators (CNOs) offer a novel adaptation of CNN architectures to learn solution operators of PDEs \cite{raonic2023convolutional}. Unlike traditional neural networks, CNOs can process functions as inputs and outputs while preserving their continuous nature, even in a discretized form. When tested on various PDE benchmarks with multi-scale solutions, CNOs outperform baseline methods significantly. This suggests that CNOs present a promising alternative framework for robust and accurate operator learning, offering advantages over SFNOs in specific applications. Compared to FNOs, CNOs minimize aliasing errors by preserving the continuous nature of functions even in discretized form, reducing artifacts. While CNO outperforms DeepONets and FNOs along most metrics, benchmarking between CNO and SFNO is yet to be performed. 

The above approaches and their key meteorological advantages and influence on ML design are summarised in Table \ref{tab:comparison}.

\begin{table}[ht]
\caption{Comparison and Proposed Approach}
\label{tab:comparison}
\begin{tabular}{p{3.5cm}p{3.5cm}p{3.5cm}}
\textbf{Approach} & \textbf{Meteo Advantages} & \textbf{ML Design Criteria} \\
\hline
Pangu-Weather \cite{bi2023accurate} & 3D Earth-specific transformer captures complex patterns in weather data with good forecast accuracy. Hierarchical temporal aggregation reduces computational costs. & Integrate 3D Earth-specific transformer with hierarchical attention \\
\hline
Cachay et al. \cite{cachay2023dyffusion} & Directly couples temporal dynamics with diffusion steps, enabling long-range forecasting. & Add diffusion for longer rollouts. \\
\hline
GCANs \cite{ruhe2023geometric} & Leverage symmetry group transformations and geometric algebras for modeling complex dynamical systems. & Potential for improved complex non-linear representations. \\
\hline
SSL LeCun \cite{mialon2023self} & SSL and joint embedding methods learn general-purpose representations of PDEs from heterogeneous data. & Explore SSL and joint embedding methods for better PDE representation and time-stepping \\
\hline
SFNOs \cite{bonev2023spherical} & Generalization of FNOs for atmospheric dynamics on spherical geometries, stable auto-regressive rollouts. & Explore spherical harmonics with Fourier transformations and operator learning for PDE representations \\
\hline
CNOs \cite{raonic2023convolutional} & Process functions as inputs/outputs, preserve continuous nature in a discretized form. & Use CNOs to minimize aliasing errors and improve operator learning of FNOs \\
\hline
\end{tabular}
\end{table}

\subsection{Meteorological Perspective on Foundation Modeling}

 There is a relatively open playground in the Earth System FM space, with technology mature and feasible enough to build a weather foundation model for the scales indicated in Figure \ref{fig:WeatherFM_meteo}. Additionally, there is scope for feasible and high-impact innovation for multi-scale weather foundation modeling.

\subsubsection{Physical Scales}\label{sec:design:scales}

As mentioned at the beginning of Section \ref{sec:design}, for effective weather forecasting, capturing information at multiple spatiotemporal scales is essential, which raises the question of whether we can train a multiscale model that can handle different resolutions and time intervals. Some deep learning models, like GraphCast \cite{lam2022graphcast}, have demonstrated the ability to forecast at different time horizons, but further research is needed to explore more complex multiscale architectures.

In the analysis outlined in Section \ref{fig:WeatherFM_meteo}, it is apparent that various scales are at play within the current forecasting approaches in the community. Presently, most AI emulators, including the likes of Pangu \cite{bi2023accurate}, FourCastNet \cite{bonev2023spherical}, and GraphCast \cite{lam2022graphcast}, predominantly operate at single spatial scales, typically favoring ERA-5 owing to the ease of working with extensive training data it provides.

Despite the considerable cost, at least two times the training cost, and likely higher due to 20 times resolution differences, a single foundation model should be able to emulate multiple datasets with MERRA2, ERA5, and HRRR as fine-tuning. This ability means the foundation model must handle different resolutions (as explained with HRRR) and time steps and provide multi-regional coverage. It is important to note that HRRR is limited to North America, while the others have global coverage.

The situation necessitates a critical decision-making process where one has to select specific scales to concentrate their efforts on, aligning with the capabilities and constraints of the current AI models. Moreover, while the possibility of modeling across multiple scales cannot be wholly discounted, it seems feasible only within a certain, limited range, as discussed in Section \ref{sec:applications:forecasting}. Currently, strategies that deploy separate supervised models for ERA5 and HRRR are necessary. However, a multi-resolution and multi-physics FM should leverage the high spatiotemporal resolution offered by HRRR and the global availability of ERA5 data. 

In summary, innovation in this area would benefit from the development of a multi-scale and multi-physics foundation model capable of efficiently approximating solutions to multiple PDEs without the necessity of stringent spatiotemporal boundary conditions that a 25x resolution increase (e.g., from ERA-5 to HRRR) would imply. Such a multi-scale and multi-physics FM could surpass existing state-of-the-art models' capabilities.

Additionally, there is an opportunity for orders of magnitude speed-up in inference vs. ECMWF IFS on multiple downstream tasks, e.g., not (only) tropical cyclones.

\subsection{Long Term Rollout}
\label{sec:design:rollout}

Understanding the long-time-range behavior of ML weather models is essential, especially for climate simulations and extreme weather prediction. AI-based forecast emulators \cite{bi2023accurate} and \cite{lam2022graphcast} have shown that their forecasts become somewhat blurry over longer time horizons. Evaluating AI-based weather models' long-term performance and stability is crucial for their practical applicability. The stability of long-term rollouts and better fidelity for extremes is a key issue that now-casting, weather, and climate foundation models would need to address. The shortcomings of all-weather AI models trained so far are the behavior on long time ranges and the ability to capture extremes.

Evidence in the literature \cite{lippe2023pde} shows that roll-out instabilities can be caused by errors in subleading modes in frequency space. This means one can achieve increased stability via diffusion \cite{lippe2023pde} or suitable Fourier space operators \cite{bonev2023spherical}. Alternatively, one can address this problem via suitable loss functions or training procedures. We will discuss this further in the Section \ref{sec:implementation:pre-training_and_loss}. SFNO \cite{bonev2023spherical} and DLWP U-Net \cite{Weyn_2020} demonstrate autoregressive 1 year rollout without instability by combating polar singularities. SFNO uses the spherical harmonic transform at the expense of spatial locality to reduce artifacts. At the same time, DLWP U-Net changes the data representation from latitude-longitude to an equiangular gnomonic cubed sphere and uses a CNN architecture to preserve spatial locality. However, autoregressive rollout without artifacts is not the same as achieving long-term forecasting stability, where the model does not simply collapse to a climatology and with blurred out extremes. The latter long-term stable rollout as a forecast with AI models is not yet competitive with physics models.

The PDE-Refiner paper \cite{lippe2023pde} studies this problem for several PDEs in one and two dimensions. As mentioned above, the authors find clear evidence that rollout instabilities arise from subleading modes that start affecting the dynamics on longer time scales. Their approach then uses refinement steps as a denoising objective of the diffusion model to shift the focus from only high-amplitude to varying amplitudes. Each refinement step adds Gaussian noise, and the model predicts this noise to denoise/refine its input.
 
Future innovations here must improve the forecast horizon gain compared to Pangu-Weather, ECMWF IFS, GraphCast, and FourCastNet. Additionally, they must extend the seven-day ahead forecasts stably to longer time scales.

\subsection{Data}\label{sec:design:data}

\subsubsection{Observational, Reanalysis, and Forecast Data}

One must decide whether to work with a gridded reanalysis product or data from observations. The latter means that the model must be able to deal with sparse and heterogeneous, often discrete and stochastic input signals from automated weather stations and satellite Earth observations. Unlike sophisticated data assimilation systems used by operational meteorological agencies, most existing AI weather forecasting models are based on gridded analysis and cannot consume observations directly.

Observations like weather station data directly measure atmospheric variables at specific locations. On the other hand, gridded data products, such as reanalysis datasets like MERRA-2 and ERA5, offer interpolated and assimilated data on a regular grid covering the entire globe. Both types of data have their advantages and limitations. Observations offer ground truth information but may have sparse coverage, especially in remote or oceanic regions. Gridded data products provide global coverage but may suffer from interpolation errors and biases introduced during assimilation. The incorporation of sparse data is currently not addressed by state-of-the-art AI models, as it would require a foundation modeling system that can perform data assimilation. 

\subsubsection{Data Volumes}

The current state-of-the-art AI weather forecasting models leveraging gridded analysis curate a fixed subset of parameters and pressure levels, sometimes also time steps. This is to some extent driven by GPU memory concerns since, e.g., global ERA5 data at all 37 pressure levels comprises 921 MB / time step (at 32-bit resolution). As noted by the authors of GraphCast \cite{lam2022graphcast}, relative data sizes per time step in the field of weather and climate are orders of magnitude higher than what most deep learning models are typically trained on, e.g., with ImageNet \cite{deng2009imagenet} and MNIST \cite{lecun1998mnist}.
The temporal resolution also varies significantly across these datasets. CMIP6 focuses on multi-decadal climate predictions, while MERRA-2 offers daily scale data for trend monitoring over the years. ERA5 and HRRR present an even finer temporal resolution, with data available hourly, which aids in various analyses, including real-time weather forecasting.

Moreover, computational demands and data volume increase progressively from CMIP6 to HRRR. While CMIP6 requires moderate computational resources, the demand escalates with MERRA-2 and ERA5, requiring more sophisticated downscaling methods and additional data integration. Transitioning to HRRR, the complexity intensifies further, necessitating high-fidelity models and substantial computational resources. Moreover, data accessibility varies across these datasets, with increasing challenges in data management and storage, especially for the high-frequency data generated by HRRR. Consequently, transitioning from CMIP6 to HRRR involves shifting from macro-level climate modeling to granular, real-time weather prediction, coupled with increasing computational and data management challenges.

The ERA5 dataset, generated by the ECMWF under the Copernicus Climate Change Service project, is a comprehensive repository of weather and climate data spanning several decades. It is updated hourly and presents a rich source for analyzing historical weather patterns and global climate trends. Users can access meteorological variables such as temperature, precipitation, wind speed and direction, sea-level pressure, and humidity. The format of this dataset is generally in a grid of $30 \times 30$ km, but for a more in-depth examination of smaller features, finer resolution data ($10 \times 10$ km) is also accessible. The latest advanced machine learning weather forecasting models frequently utilize this dataset, as illustrated in Figure \ref{fig:WeatherFM_meteo}.

The High-Resolution Rapid Refresh (HRRR) dataset is produced by the highest resolution forecast model operated by the National Center of Environmental Prediction (NCEP) for the North American region . The HRRR model utilizes a grid with horizontal cell dimensions of $3 \times 3$ km cells and updates hourly to incorporate the latest observations. HRRR uses the Weather Research and Forecasting (WRF) model assimilating various observations, such as radar data, satellite retrievals, and surface observations to produce high-fidelity short-term forecasts. Its high temporal update frequency allows for observing and predicting rapid weather developments, making it suitable for monitoring and forecasting severe weather events, including thunderstorms and tornadoes.

\subsection{Diagnostics}

To evaluate the proposed approaches and models from a design perspective, we can further seek to answer the following diagnostic questions: How can we evaluate architecture choices independent of downstream tasks? Is there a set of criteria to evaluate at the pre-training time that suggests good downstream performance? These diagnostics can help compare an AI weather FM model against physics models like ECMWF IFS, operational models like HRES trained with reanalysis datasets like ERA-5 and MERRA-2, and high-resolution regional models like HRRR.

\subsubsection{Architecture Evaluation}
 \begin{enumerate} 
 
 \item \textbf{Meteorological relevance}: Assess the proposed models' architecture choices regarding their relevance to meteorological processes. Verify if the models effectively capture the vertical structure of the atmosphere, handle complex weather patterns, and encode important meteorological features.
 \item \textbf{Grid representation}: Evaluate the model's ability to work with meteorological data on a global scale and adapt to different spatial resolutions commonly used in meteorological simulations.
 \item \textbf{Interpretable features}: Investigate whether the model can generate interpretable features that align with known meteorological phenomena, such as cyclones, fronts, and atmospheric stability.
 \item \textbf{Spherical vs. Cartesian grid representations:} The choice of grid representation in Earth system models using machine learning profoundly impacts model performance and interpretation, whether spherical, Cartesian, or other. Spherical grids align with Earth's geometry, enhancing accuracy for global phenomena and ensuring better physical consistency, especially at the poles. In contrast, Cartesian grids offer computational simplicity and are more suitable for regional models, though they can introduce errors in large-scale applications. Physical principles, like conservation of mass and energy, maybe better upheld in global, spherical representations. Machine learning models might require different feature engineering based on the chosen grid. Transfer learning between grids is challenging due to data representation differences. Other grids, like hexagonal or icosahedral \cite{bonev2023spherical} \cite{bi2023accurate} \cite{lam2022graphcast}, strike a balance by avoiding issues like pole singularities present in lat-lon grids. Different grids can better represent specific phenomena; for instance, large-scale circulations fit spherical grids, while localized events might favor Cartesian grids. The optimal grid choice hinges on application, scale, and desired model properties. Grid selection is pivotal for accuracy, efficiency, and consistency in Earth system models integrating machine learning. 
 \end{enumerate}

\subsubsection{Pre-training Evaluation}
 \begin{enumerate} 
 
 \item \textbf{Transfer learning}: Analyze how well the pre-trained models transfer knowledge to downstream meteorological applications. Evaluate their performance on various meteorological prediction tasks, such as temperature, precipitation, wind, and pressure forecasting.
 \item \textbf{Robustness}: Examine the models' robustness to different climate regions, extreme weather events, and data outliers commonly encountered in meteorological datasets.
 \item \textbf{Time-scale dependency}: Investigate how the pre-trained models handle short-term and long-term weather forecasts. Assess their ability to capture daily variations and seasonal changes.
 \item \textbf{Forecasting vs.~generic pretext tasks:} While weather forecasting is the ultimate goal, researchers have also explored using generic pretext tasks for pre-training meteorological models. Pretext tasks, such as contrastive learning and masking, help learn general-purpose representations from unlabelled data, which can be transferred to downstream forecasting tasks \cite{assran2023self,he2022masked,hoffmann2023atmodist,nguyen2023climax}. The effectiveness of these pretext tasks in improving weather forecasting capabilities requires extensive evaluation and comparison against traditional forecasting approaches.
 \end{enumerate}
 
 \subsubsection{Comparison Against Physics Benchmark}

 \begin{enumerate}
 \item \textbf{Accuracy and performance}: Compare the proposed models' forecasting accuracy against the ECMWF IFS model, a widely used benchmark for global weather prediction. Evaluate their skill in representing meteorological variables over different forecast lead times.
 \item \textbf{Computational efficiency}: Assess the proposed models' computational efficiency compared to IFS ECMWF, which is crucial for real-time weather forecasting applications.
 \item \textbf{Data representation}: Verify if the proposed models can effectively represent the unique characteristics of MERRA-2 and ERA5 datasets, which may vary in data quality, spatial resolution, and temporal coverage.
 \item \textbf{Generalization}: Evaluate the models' ability to generalize across different datasets and climate regions, mainly for areas with limited observation data.
 \item \textbf{Information flow across arbitrary length and time scales:} In weather forecasting, information flow across arbitrary length and time scales is crucial for capturing the full dynamics of the atmosphere. Models like the Swin Transformer \cite{liu2021swin} have shown promise in handling long-range dependencies efficiently. This is less relevant in the case of a weather FM, where the effects of teleconnections and oscillations are explicit in the initial conditions and temporal variability over forecast time scales are negligible.
 \end{enumerate}

These evaluations and comparisons enable us to pinpoint the most appropriate strategy for pre-training foundational AI models from a meteorology and climate science perspective. The selected strategy should effectively and robustly encode both the intrinsic behaviors and processes occurring in the Earth's atmosphere (atmospheric dynamics) as well as the computational methods and algorithms used to simulate these behaviors and processes in weather models (model dynamics). This would ideally enhance weather forecasting capabilities, potentially surpassing conventional NWP methods like IFS ECMWF by facilitating broader generalization across various scales.

\subsection{Impact vs Effort}

\subsubsection{Key Characteristics for ML Architecture} \label{sec:design:keys}

In summary, from the above diagnostics, we recommend that a WeatherFM must consider the following key characteristics for ML architecture implementation in the subsequent Section \ref{sec:implementation}:

 \begin{enumerate}
 \item Multi-scale modeling
 \item Stability of long-term rollout
 \item FMs' ability to add value (in lowering complexity and resources for fine-tuning)
 \item Data availability considering constraints with sparse inputs 
 \end{enumerate}
 
 From the prior sections, we recommend that the community focus on spatial and temporal scales related to a WeatherFM; a complete Earth system FM is out of scope. Beyond the spatial scales recommended in Figure 1, dealing with microscale regimes using nowcasting methods between seconds to less than 1 hour would require significant effort. Microscale scales can destabilize the proposed weather FM, impacting the downstream tasks in the other spatiotemporal regimes.
 
 In addition to super-resolving microscale, another innovation that nowcasting would necessitate is the inclusion of data assimilation from sparse observations into a foundation model. While this is feasible, it could require an extended project effort beyond the first year focusing on this topic after gaining experience building a short-term forecasting weather foundation model. 
 Therefore, in the subsequent section, we consider less than 1 hour and less than 1 km resolutions for short-term weather forecasting, where we have greater confidence in innovations with high impact.

 Beyond 14 days, with medium-term weather and sub-seasonal forecasting, machine learning and physics-based models have yet to demonstrate significantly better results than climatologies. These regimes would require significant innovations with high uncertainty.
 
 \paragraph{Future Opportunities for Innovation Beyond the Weather Foundation Model}
 
 When considering scales extending to multi-year, long-term forecasts, two fundamental aspects come to the forefront: climate prediction and climate projections. The former relates to modeling and forecasting natural phenomena like global climate oscillations, including ENSO, using extensive resources such as the IPCC CMIP6 datasets.

 On the other hand, climate projections delve deeper, involving the consideration of potential scenarios concerning human responses to climate crises. This necessitates the development of machine learning models capable of assimilating how alterations in specific parameters, such as global CO2 concentrations, might influence the climate's evolution over time.

 However, a significant obstacle remains in the prediction of complex phenomena like global climate oscillations, including ENSO, and understanding the uncertainties revolving around the changing dynamics of the global carbon cycle and Scope 1-3 emissions. An added layer of complexity is brought about by the uncertainties linked with human reactions to climate crises and the associated feedback and tipping points. Despite the challenges, there lies a vast arena for innovation, as climate scientists continue to navigate these intricate questions with the objective of fostering more accurate and encompassing climate predictions and projections.

 Another area for improvement with this range is that one can only back-test predictions and would need to wait several years to ensure its impact when data is available. However, this may not be a constraint for adoption as a tool for scenario analysis with sufficient back-testing with climate change models. There is an opportunity to speed up models performing hundreds of years of simulations and increase their multi-scale fidelity.

\section{Implementation}
\label{sec:implementation}

This section discusses concrete requirements for implementing a WeatherFM and compares the tradeoffs between using different components in the model, such as transformer blocks, data representations, and neural operators.

As discussed in Section~\ref{sec:design:keys}, attending to two key objectives is crucial when developing a weather model: multi-scale robustness and long-term stability. The first enables the model to effectively handle diverse data sources that may vary in distance and time. Meanwhile, long-term stability guarantees the model's dependability even when the lead time exceeds initial considerations during its design. Finally, it is essential to address any gaps in the code that may affect computational performance when dealing with the vast amounts of data involved in weather and climate.

\subsection{Data Representation}
\label{sec:implementation:data_representation}

Data representation is the first step and is a crucial component when designing any machine-learning solution. While weather and climate data may share some similarities with image and video data, the utilization of off-the-shelf systems may prove inadequate due to the unique characteristics of the former. This section discusses some important considerations regarding weather and climate data.

\subsubsection{Coordinate Representation} \label{sec:implementation:coordinate}

There are obvious reasons to utilize a Cartesian grid to represent weather and climate data. Such data share common characteristics with 2D/3D images and videos that are abundant on the internet, and their applications are much more popular. When we look at the literature on applying machine learning to weather and climate data, their bases are usually from work like U-Net \cite{ronneberger2015u}, ResNet \cite{he2016deep}, ViT \cite{dosovitskiy2020an}, and VideoMAE \cite{tong2022videomae}, which were initially intended for images and videos.

Usually, these models use climate data in a uniform rectangular gridded pattern. Some examples include ClimaX \cite{nguyen2023climax}, Pangu \cite{bi2023accurate}, SwinRDM \cite{chen2023swinrdm}, DYffusion \cite{cachay2023dyffusion}, FourCastNet \cite{pathak2022fourcastnet}, and MetNet-2 \cite{espeholt2022deep}. While this approach may be suitable for local and regional modeling attributes, it fails to capture the physical and geodesic properties of the Earth, especially as the focus shifts away from the equator \cite{cachay2022climformer}.

In another direction, a few works explore spherical representations such as equirectangular (ERP) and cubes \cite{cho2022using}, icosahedral grids \cite{cachay2022climformer, lam2022graphcast, cho2022using} and the Hierarchical Equal Area iso-Latitude Pixelation (HEALPix) \cite{gorski2005healpix}. In \cite{bonev2023spherical}, the authors discuss a new version of Fourier neural operators for the sphere. They introduce the concept of spherical convolution to make the Fourier layer linear and equivariant. HEAL-SWIN \cite{carlsson2023heal} presents a combination of the HEALPix pixelation of the sphere with the Swin transformer's hierarchical structure. HEALPix is especially suitable for this approach since it defines nested pixelation. Unfortunately, there seems to be no argument at present that establishes the impact of the additional engineering work when moving from a Cartesian grid to a spherical representation. The decision has to be made via ablation studies.

\subsection{Model Components}
\label{sec:implementation:components}

In the following sections, we will discuss the main candidates for building blocks for foundation models for weather and climate. These are transformers, graph neural networks, and different forms of neural operators. Most existing publications on large AI models for weather focus on one of these. Yet, in principle, they can also be combined.

\subsubsection{Transformers}
\label{sec:model:transformer}

To overcome the bottleneck problem with fixed-length encoding vectors, the authors in \cite{bahdanau2014neural} introduced the attention mechanism. This method is especially crucial for longer or more intricate sequences, as their representation would otherwise be confined to the same dimensionality as shorter or simpler sequences. As an extension of Bahdanau's work, the Transformer model replaces the original attention with self-attention that comprises a Query-Key-Value ($QKV$) mechanism. Equation~\ref{eq:att} gives the self-attention used in the transformer model for the given matrix representations of queries $\textbf{Q} \in \mathbb{R}^{N \times D_k}$, keys $\textbf{K} \in \mathbb{R}^{M \times D_k}$, and values $\textbf{V} \in \mathbb{R}^{M\times D_v}$.
\begin{equation}\label{eq:att}
 \text{Attention}(\textbf{Q}, \textbf{K}, \textbf{V})=\text{softmax} \left(\frac{\textbf{Q}\textbf{K}^T}{\sqrt{D_k}}\right) \textbf{V}=\textbf{AV},
\end{equation}
where $N$ represents the length of queries, and $M$ represents the length of keys (or values). $D_k$ and $D_v$ represent the dimensions of keys (or queries) and values, respectively. $\textbf{A}=\text{softmax} \left(\frac{\textbf{Q}\textbf{K}^T}{\sqrt{D_k}}\right)$ is the attention matrix, and softmax is applied row-wise. The dot-products of queries and keys are divided by $\sqrt{D_k}$ to address the issue of the gradient vanishing. Transformer uses multi-head attention instead of a single attention function. The original queries, keys, and values with $D_m$ dimensions are projected into $D_k$, $D_k$, and $D_v$ dimensions with H different sets of learned projections.

Current foundation models in natural language, image, video, and time-series processing rely on Transformers and represent the state-of-the-art in each area (AI Index, 2023). Also, Transformers already show promising results in weather forecasting, as mentioned before. Hence, the possibility of creating a foundational model for weather and climate based on Transformers is an exciting prospect.

\paragraph{Tokenization Strategies}

Transformers consume data in discrete chunks, known as tokens. Converting raw input data, such as text, images, or time series, into tokens is referred to as tokenization. The process is relatively straightforward for language since words are natural units of information. There is no natural definition for other types of data, and different choices heavily affect model performance.

After breaking the input data into tokens, it is necessary to embed them into a metric space. As we discussed, weather and climate variables have similarities with 3D videos. However, atmospheric data tends to have more variables (channels) with different correlations. For instance, temperature and humidity interact differently compared to wind and geopotential. In Pangu \cite{bi2023accurate}, variables are encoded as channels in a 3D convolution, which results in different variables with different meanings sharing the same filters.

On the other hand, ClimaX \cite{nguyen2023climax} has separate patching for each variable, but it leads to increased model complexity. Of course, Pangu's choice reduces the complexity of the model and makes it faster for training and serving. However, such a choice limits the model to become a foundation model since downstream tasks may not have all the variables used during training, which would make the model fail.

A foundation model for climate and weather must be flexible to deal with multiple input configurations, including different sets of variables or resolutions, but without increasing the complexity too much. Also, the tokenization may need to be robust when there is a sparse input, the data is 2D or 3D, or the resolution varies. The tokenization scheme may better explain model outputs if it highlights a good vocabulary (``climate words''), which could be identifiable patterns for atmospheric variables. Finally, the tokenization may allow the model to learn long-range connections in space and time, for example, teleconnections.

\paragraph{Positional Encoding}

After tokenization, it is best to add a signal to each token that informs the model about the origin of the token. This is referred to as positional encoding. This signal can indicate absolute or relative position, but it can also discriminate between different physical parameters or time steps. In what follows, we will discuss three key properties.

First, positional tokens help to directly inform the model about the content of the remaining tokens. With positional tokens, we can inform the model on the unit of information (e.g., ``the embedded tokens refer to humidity at 100hPa'') or use positional tokens to convey information on the topography (e.g., ``the current area of interest is over land, ocean, city, ...''). Especially in multi-regional, multi-modal settings, positional tokens help us to inform the model of differences of the embedded input tokens.

Second, positional tokens can be hard-coded (positional encoding) or learned (positional embedding). We distinguish between absolute, relative, or conditional encodings for positional encodings. Absolute positional encodings have the same dimension as the input tokens' embeddings and refer deterministically to a specific index in the data, for example, the latitude. Relative positional encodings consider the distance between the tokens, and conditional positional encodings are generated dynamically and dependent on the surrounding area of the input tokens. Positional embeddings offer another approach to representing positions: learning a set of positional tokens for each position \cite{gehring2017convolutional, devlin2018bert}. Learning embeddings provide greater flexibility than encoded position representation as they can be adapted to specific tasks through back-propagation.

Third, we can leverage positional embeddings to generalize across pressure levels during inference. During training, data is available for a few specific pressure levels. When considering the positional embedding as a function of the pressure level, we can generalize to different levels during inference unseen during training.

Overall, positional tokens offer to encode or embed additional information, handle arbitrary sequence lengths, and generalize across pressure levels during inference. These properties can be beneficial in many weather modeling scenarios and are specific to the transformer architecture---compared to, for example, graph-based approaches. Recent transformers for weather forecasting have, for example, employed learnable absolute positional tokens of the earth's surface to embed information on variables and the location of values \cite{bi2023accurate}. 

\paragraph{Sequence Length}

Handling arbitrary sequence length is an important prerequisite in weather modeling. Handling sequences longer than sequences seen in the training time (i.e., ``inductive'' approaches) is particularly relevant. For example, when training a transformer for hourly prediction on ERA5 reanalysis data of the highest resolution (i.e., 0.25° $\times$ 0.25°), we have a 2D input resolution of 1440 $\times$ 721, with an additional 37 pressure levels for the third dimension. When we consider humidity, temperature, geopotential, easterly, northerly, and vertical wind velocities, we obtain a total of 1440 $\times$ 721 $\times$ 37 $\times$ 6 = 230,489,280 data points at each single time step (see above, c.f. \cite{bi2023accurate}). Sequence lengths are typically relatively short for model training to keep the associated computational efforts manageable (e.g., hundreds of thousands to a few million tokens per sequence in computer vision or natural language processing). This would refer to a 2D spatial resolution of approximately 100 $\times$ 100 (e.g., 100 $\times$ 100 $\times$ 37 $\times$ 6 = 2,220,000). However, for long-term forecasts or global forecasts, models have to be able to digest more data---i.e., longer sequences. It is possible to achieve such an ability with positional embeddings (i.e., learned positional tokens) and relative and conditional positional encodings (i.e., hand-crafted positional tokens).

\paragraph{Sparse Attention}

The attention mechanism described earlier allows each query node to gather information from all memory nodes, creating a fully connected bipartite graph. While this approach can improve representation, the size of matrix $A$ in Equation \eqref{eq:att} can become huge depending on the input and significantly impact complexity and memory usage. Conversely, the sparse attention model eliminates specific connections between nodes $A_{ij} = q_ik_j^T$, yielding less memory usage and lower complexity. Also, it is possible to add structural bias in this process.

The representation of atmospheric variables occurs in a 3D space. The finite speed of the fluid motion means that it takes time for one point in space to affect another. Therefore, the model does not need to focus on all positions at all times, thus reducing the number of $k_i$ and $q_j^T$ to attend. One example of how the model can reduce resource usage is by utilizing a band \cite{guo2019star} or dilated \cite{oord2018representation} attention while improving its learning capabilities. Furthermore, depending on how redundant is the input signal, one can disregard part of the input space and apply random sparse attention, similar to Big Bird \cite{zaheer2020big}.

\subsubsection{Graphs}

In Section~\ref{sec:implementation:data_representation}, we explored the typical representation of data in Transformer models, which resembles dense tensors, e.g., images. Additionally, we noted that the typical attention mechanism could be viewed as a fully connected bipartite graph. In this subsection, we will discuss the possibility of directly using a graph structure to represent weather data and the flow of information between different pixels and regions.

Graph Neural Networks (GNNs) have emerged as powerful tools in weather forecasting. Graph Convolutional Networks (GCNs), Graph Attention Networks (GATs), Graph Autoencoders (GAEs), and Graph Diffusion Models have proven effective in learning representations, predicting spatial distributions of meteorological elements, focusing on specific nodes, and deriving latent representations. Ongoing research, such as \cite{lam2022graphcast, chen2022resgraphnet, wu2019graphwavenet, ying2018hierarchical}, indicates that graph-based methodologies have even more innovative and efficacious applications in weather forecasting. Furthermore, recent work on GNN with new architecture with message passing showed great results in solving partial differential equations over meshes \cite{sanchez2020learning, pfaff2020learning}. This work proposed an Encoder-Processor-Decoder architecture, where the encoder encodes the existing mesh or grid into multigraph, processor uses $N$ identical message passing blocks, which generalize GraphNet blocks to multiple edge sets followed by a decoder that predicts the state $t+1$ from input timestep $T$ and decode the transformed mesh to original space.

Note that a critical characteristic of a weather foundation model is the flexibility regarding different input configurations, as discussed in Section~\ref{sec:implementation:data_representation}. Like approaches that encode multiple variables in one convolution, GNN encoders may require retraining for a different input set, affecting their ability to run multiple downstream tasks.

\subsubsection{Neural Operators}

Over the past few years, scientists and engineers have struggled with complex calculations and need more data specificity in traditional methods for solving partial differential equations (PDEs), which describe various phenomena in different fields. Standard numerical methods like finite difference and finite element techniques are known for being computationally taxing, especially in high-dimensional situations, making finding efficient solutions challenging.

Data-driven approaches like DeepONets \cite{lu2021learning} and Fourier Neural Operator (FNO) \cite{li2020fourier} are promising for solving PDEs as they can learn operators directly from data. They are efficient in handling complex operator learning tasks and can generalize well. Two models that are particularly compelling for weather forecasting are FourCastNet and SFNO. FourCastNet combines the Adaptive Fourier Neural Operator (AFNO) model with a transformer, while SFNO generalizes FNOs on the sphere to learn operators on spherical geometries. In particular, the FourCastNet model utilizes the attention mechanism with AFNO for efficient token mixing with a ViT backbone. SFNO is based on a generalized Riemannian manifold, in this case a sphere. 

One particular approach seems more efficient and robust for integrating with transformers in weather prediction: the Convolutional Neural Operator (CNO) \cite{raonic2023convolutional}. Section˜\ref{sec:model:transformer} discusses tokenization as an essential step when using a Transformer backbone. For imagery, convolution layers are the first choice to accomplish this. However, CNNs typically have finite-dimensional input-output mappings and have shown dependency on grid resolutions when directly applied to PDE problems. By introducing Convolutional Neural Operators (CNOs), the authors overcome these barriers, creating a system capable of learning operators that can map input functions to output functions.

CNOs are not strictly bound by the limitations of grid resolutions, which have been a significant bottleneck for other networks. By mapping input functions directly to output functions, they can handle diverse datasets without specific grid dependencies. Such feature makes them more adaptable and flexible, especially regarding weather data. Also, Transformers, known for their parallelization and attention mechanisms, can integrate seamlessly with CNOs, creating a synergy that leverages the strengths of both architectures. This integration can lead to models that can handle the high-dimensional data prevalent in weather prediction with more skill and accuracy compared to the integration with other networks like DeepONets and FourCastNet.

\label{sec:implementation:pre-training_and_loss}
\subsection{Pretraining Regimes and Loss Functions}

Having discussed possible choices of data representations and model backbones in Sections \ref{sec:implementation:data_representation} and \ref{sec:implementation:components} respectively, it is now time to turn to the final ingredient when implementing a foundation model: the choice of a pre-training regime.

\subsubsection{RMSE-Type Losses: Reconstruction and Forecasting}

Throughout this paper, we have repeatedly alluded to the similarities between gridded weather and climate data on the one hand and images or videos on the other. In the context of pre-training, the current standard approach is that of masking \cite{he2022masked}. When training a vision transformer, one removes a random subset of tokens. The remaining tokens are encoded. The decoder is given the latent space representation as well as information about the missing tokens. The task is then to reconstruct the entire image. Typically this is trained with an RMSE loss.

For weather and climate data, one re-phrase a variety of pretext tasks as variations of many variations of the theme of masking. E.g.~the authors of \cite{nguyen2023climax} use forecasting as a pretext task. That is, given some input at time $t$ the model is asked to generate the same atmospheric fields at time $t+\Delta t$. Instead of informing the decoder about the spatial location of the missing tiles, the decoder is handed a temporal position encoding that reflects $\Delta t$. While not implemented in the literature, one could naturally use a similar pattern to remove entire fields or pressure levels from the input and reconstruct those. We can think of all these cases -- even the forecasting one -- as reconstructing missing information by optimizing an RMSE loss.

The recently published and transformer-based AtmoRep \cite{lessig2023atmorep} also uses masking. However, reconstruction is not deterministic and or trained by a simple pixel-wise RMSE loss. Instead, AtmoRep generates an ensemble of predictions that allows for a probabilistic treatment. The loss function is custom and involves the first two statistical moments of the predicted ensemble.

\subsubsection{Alternative Approaches}

\paragraph{Contrastive Learning}

The key property of the reconstruction tasks mentioned above is that losses are computed with pixel-by-pixel comparison. For certain situations, this might not be ideal. One example would be sparse fields like precipitation or snowfall. One well-established alternative to pixel-wise losses is contrastive learning. Given data input $x$, a model $f$, and a set of randomized transformations $T$, the SimCLR approach of \cite{chen2020simple} is to create two augmentations $x' = T(x)$ and $x'' = T(x)$. The model $f$ is then trained such that $f(x')$ and $f(x'')$ are close in latent space yet far from other data. This is achieved by optimizing the ``Normalized Temperature-scaled Cross Entropy'' (NT-Xent).\footnote{To be more precise, the model $f$ has two components: $f = h \circ g$. $g$ embeds data in latent space. $h$ is an additional projection that reduces the dimensionality of the embedding vector further. The optimization target is defined on the projected space of $h\circ g(x)$. Yet after training, $h$ is discarded and one uses latent space embedding vectors $g(x)$.} To ensure that embeddings from dissimilar data are well separated in latent space, NT-Xent involves all samples from the current batch when training the model. In the case of \cite{chen2020simple} this leads to huge memory and resource requirements since batch sizes can be as large as 4,096. However, there are improvements on the original paper, such as \cite{he2020momentum} that manage to circumvent this issue.

A key aspect of contrastive approaches is the design and selection of the transformations $T$. Common choices are cropping, blurring, masking, color jitter, and Sobel filters. There were efforts to define custom transformations that are specific to atmospheric dynamics. However, it was shown in \cite{nathaniel2023resource} that at least for weather and climate there is no need for complicated or custom pretext tasks. Instead, it is entirely sufficient to use sub-sampling as the sole augmentation: Given a data cube $x$ that has been tokenized Section \ref{sec:model:transformer}, one simply samples a small subset of tokens and discards the rest. This is reminiscent of masking, yet it bypasses the need to hold the entire data cube in memory. Another recent approach that does not rely on hand-crafted data augmentations is the Image-based Joint-Embedding Predictive Architecture (I-JEPA) \cite{assran2023self}.

\paragraph{Domain-Specific Pretext Tasks}

While pre-training is meant to be task-independent, it does not have to be domain-independent. In other words, while the training regimes discussed so far were originally developed in computer vision, we will now turn to approaches specific to weather and climate data.

A prime example of this is AtmoDist \cite{hoffmann2023atmodist}. The authors propose a novel self-supervised learning strategy for representation learning in atmospheric dynamics based on a domain-specific pretext task drawn from the principles of geophysical fluid dynamics. This task requires the model to predict temporal separation, defined as the number of time steps between two nearby atmospheric states within a given temporal sequence of these states. The efficacy of the learning is evaluated using the categorical cross-entropy loss function, which measures the discrepancy between the predicted and actual temporal separations. Additionally, the paper suggests other potential domain-specific pretext tasks for representation learning in this field, which include predicting spatial separation, temporal evolution, inter-state relationships, and the effects of external forces on atmospheric states.

\subsubsection{Diffusion}

A standard probabilistic diffusion regime consists of a forward noising process and a reverse denoising process to learn the data distribution. With respect to weather forecasting, diffusion is used for probabilistic forecasting in \cite{cachay2023dyffusion} \cite{li2023seeds} and long-term rollouts in \cite{cachay2023dyffusion} \cite{lippe2023pde}.

As shown in \cite{li2023seeds} \cite{cachay2023dyffusion} diffusion models have the ability to generate probabilistic forecasts. Dyfussion \cite{cachay2023dyffusion} uses a forecasting objective instead of the standard denoising objective. The interpolator creates stochastic outputs using Monte Carlo dropout to generate probabilistic forecasts during the reverse process. SEEDS \cite{li2023seeds} uses 128 trained ensemble emulators to generate probabilistic forecasts that resemble weather states of the ``seeds'' (two forecasts of an event generated by two different ensemble models) provided during inference.

Error propagation in auto-regressive models makes long-term rollouts unstable and inaccurate. To improve forecasts over long-term horizons, Dyfussion \cite{cachay2023dyffusion} uses clean and noised initial conditions. PDE-Refiner \cite{lippe2023pde} uses the diffusion process to improve the predictions by looking at them iteratively to capture the often neglected low amplitude information in data.

In \cite{cachay2023dyffusion} \cite{ovadia2023ditto} diffusion inspires time-conditioned interpolation to leverage the temporal dynamics in data. Using only the forward process of diffusion models, DiTTO \cite{ovadia2023ditto} creates a continuous interpolation between the initial and final time steps. It uses time evolution instead of incremental noise in the forward process. A similar idea is used in Dyffusion \cite{cachay2023dyffusion} for the forward process to create a time-conditioned interpolater for intermediate dynamics steps.

When using diffusion models, we need to consider 
\begin{itemize}
 \item Task-specific noise strategy. For example, \cite{lippe2023pde} uses an exponentially decreasing noise scheduler with a very low minimum noise variance decreasing faster than regular diffusion models to meet the low error standards for deterministic solutions. Based on empirical evidence \cite{cachay2023dyffusion} uses a cosine (linear) scheduler for SST forecasting to map from diffusion to interpolation time step
 \item A scoring function for denoising objectives. Various scoring functions used in literature are ViT \cite{li2023seeds}, U-Net \cite{lippe2023pde} \cite{cachay2023dyffusion}, CNN \cite{cachay2023dyffusion}, UNet-Transformer \cite{ovadia2023ditto}. For our model, the choice of the backbone will determine this function.
\end{itemize} 

While diffusion models have the above-mentioned benefits, the trade-off often lies between computation resources and forecast accuracy.

\section{Summary and Conclusions}

Driven by the massive progress in the area of large language models and catalyzed by the impact of the associated chatbots beyond the confines of academia, the last year has seen the adoption of the foundation model paradigm beyond the domain of NLP. At the same time, efforts to tackle numerical weather prediction using large, supervised AI models has reached a level where it is garnering attention from the meteorological community \cite{lam2022graphcast, pathak2022fourcastnet, bi2023accurate, chen2023swinrdm}. The almost logical intersection of these two trends is the development of foundation models for weather and climate, which started in \cite{nguyen2023climax, lessig2023atmorep}. As we discussed in Section \ref{sec:applications:summary}, foundation models bring a range of benefits to weather and climate. Most notable among these are efficiency gains when comparing with conventional approaches yet also a reduction of the reliance on labels.

Despite the notable successes and the promise of the approach, the development is in its infancy. There is no clearly established or dominant paradigm for the design principles that underlie such a model nor a clear understanding about its most prominent applications and their impacts. Indeed, while both \cite{nguyen2023climax} and \cite{lessig2023atmorep} are transformer-based architectures, the forecast emulators of \cite{lam2022graphcast, pathak2022fourcastnet, bi2023accurate, chen2023swinrdm} leverage transformers, graph neural networks and neural operators. Given the considerable computational cost involved in foundation model training, one cannot make architecture choices solely based on experimentation. With this in mind, our work aims to give some guidelines to anyone interested in leveraging, designing or implementing foundation models for weather and climate domain.

Overall, our approach is straightforward yet methodologically structured: starting from possible applications (see Section \ref{sec:applications}), we aim to identify the gap and scope of a foundation model design. Given a fixed scope, meteorological concerns do impose certain requirements on foundation model design (Section \ref{sec:design}). Finally, one can use these constraints to choose suitable building blocks, data sets, pre-training procedures and  pre-training procedures, and fine-tuning strategies (Section \ref{sec:implementation}).

Following this path, we have come to several recommendations. To start, it should be clear that a current generation foundation model will have to be limited in scope regarding spatial and temporal scales.
Having said that, given that we want to capture as many downstream applications as possible, it is beneficial for a single model to cover a range of both spatial and temporal scales. Regarding concrete applications, forecasting and downscaling emerge almost due to their proximity to predominant pre-training paradigms; yet many other applications can easily be brought into scope.

However, the choice of scale(s) is not only driven by the desired downstream tasks, but also by available pre-training data. Here, it should be clear that a longer time series will capture a richer and wider set of meteorological phenomena. A straightforward way to achieve this is to train with the 80 years of available ERA5 data. Yet this anchors the scale of the model to that of ERA5. Alternative approaches would lever radar observations for a nowcast model or CMIP6 or similar for something oriented towards climate scales. Anchored with ERA5 data, a model might presumably extend all the way to the km scales of regional forecast models such as RDPS or HRRR.

The considerations of the previous paragraph point towards an AI system that is able to use gridded NWP output of varying resolution and with possibly varying parameter definitions as input. In addition, working with both ERA5 and HRRR means also taking varying spatial coverage into account. If one does make these choices, there are immediate consequences for implementation choices. To start, flexibility regarding grids as well as spatial coverage strongly suggests transformers rather than graphs. While one might argue that the breaking up of spatial information into chunks during tokenization is somewhat unnatural, the process yields considerable flexibility since one can define the concrete meaning of each token regarding spatial coverage, position and resolution via position encodings. At the same time, this means making clear plans to deal with sequence length for attention calculations. Our discussion outlined possible choices.

Finally, while forecasting is far from the only downstream task of interest, it is probably the one with the most clearly established benchmarks as well as clear impacts. Yet as soon as forecasting is in scope, one has to deal with the question of rollout stability as well as fidelity towards extremes. Again our discussion highlighted several choices.

\printbibliography

\end{document}